# Artificial intelligence supported anemia control system (AISACS) to prevent anemia in maintenance hemodialysis patients

**Toshiaki Ohara**[1,2,*], **Hiroshi Ikeda**[3], **Yoshiki Sugitani**[4], **Hiroshi Suito**[4], **Viet Quang Huy Huynh**[4], **Masaru Kinomura**[5], **Soichiro Haraguchi**[6] and **Kazufumi Sakurama**[2,7]

[1] Department of Pathology & Experimental Medicine, Okayama University Graduate School of Medicine, Dentistry and Pharmaceutical Sciences, Okayama, Japan; t_ohara@cc.okayama-u.ac.jp
[2] Department of Gastroenterological Surgery, Okayama University Graduate School of Medicine, Dentistry and Pharmaceutical Sciences, Okayama, Japan
[3] Department of Internal Medicine, Shigei Medical Research Hospital, Okayama, Japan
[4] Advanced Institute for Materials Research, Tohoku University, Miyagi, Japan
[5] Division of Hemodialysis and Apheresis, Okayama University Hospital, Okayama, Japan
[6] Kobayashi Medicine Clinic, Okayama, Japan
[7] Department of Dialysis Access Center, Shigei Medical Research Hospital, Okayama, Japan
[*] Correspondence: t_ohara@cc.okayama-u.ac.jp; Tel.: +81-86-235-7143

**Abstract:** Anemia, for which erythropoiesis-stimulating agents (ESAs) and iron supplements (ISs) are used as preventive measures, presents important difficulties for hemodialysis patients. Nevertheless, the number of physicians able to manage such medications appropriately is not keeping pace with the rapid increase of hemodialysis patients. Moreover, the high cost of ESAs imposes heavy burdens on medical insurance systems. An artificial-intelligence-supported anemia control system (AISACS) trained using administration direction data from experienced physicians has been developed by the authors. For the system, appropriate data selection and rectification techniques play important roles. Decision making related to ESAs poses a multi-class classification problem for which a two-step classification technique is introduced. Several validations have demonstrated that AISACS exhibits high performance with correct classification rates of 72%–87% and clinically appropriate classification rates of 92%–98%.

## 1. Introduction

Anemia, a common complication associated with chronic kidney disease (CKD), is a risk factor for high mortality [1]. Erythropoiesis-stimulating agents (ESAs) and iron supplements (ISs) are usually administered during hemodialysis treatment to patients. Generally, patients with large hemoglobin (Hb) variations are likely to have complications and often need to be hospitalized, and vice versa [2]. Therefore, physicians are trying to stabilize patients' Hb values within a certain range. However, doing so is very difficult because of complicated disorders such as altered iron metabolism, poor response to ESAs, and residual blood in dialysis equipment, which are mostly common problems for hemodialysis patients. Moreover, general situations such as concomitant diseases and differing backgrounds of patients in different countries [3,4] are also affecting the difficulty. Compounding these difficulties are economics concerns such as high costs of ESAs, which are heavily burdening medical insurance systems [5,6].

Although hemodialysis patients are becoming increasingly numerous worldwide, physicians who are able to manage and administer treatment appropriately are not being trained in sufficient numbers to keep pace with the increasing numbers of patients requiring hemodialysis treatment [6]. To reduce burdens on physicians and medical insurance systems under these circumstances, effective decision-making support systems are urgently anticipated. Recently, artificial intelligence (AI) technologies have been used extensively in nephrology [7,8]. Several studies conducted to assess hemodialysis have predicted vital reactions including studies specifically examining anemia control [9–12]. Model predictive control (MPC) approach was utilized and extended for effective anemia control [10-12]. Systems using AI for predicting Hb values for hemodialysis patients were presented in the literature [13,14]. Anemia control model (ACM) achieved improved control accuracy and decreased patients' need for ESAs [15,16].

Although anemia control assisted by AI technologies appears promising, a discrepancy persists between technologies and actual medical practice. Widely diverse health conditions of actual patients and various legal and economic constraints can cause many difficulties. As a result, available datasets including data of similar patients are usually not so large. Therefore, a different approach was adopted for AI learning in this study: the AI learns based on decisions of experienced physicians rather than data showing reactions of the patients' living bodies, such as Hb values. From highly experienced physicians with work histories including blood examination, we gathered data of their dosage direction decisions for patients there. To enhance the learning process, we constructed procedures for the rectification of clinical data. Then we developed an artificial-intelligence-supported anemia control system (AISCAS).

## 2. Materials and Methods

*2.1 Patients and datasets*

2.1.1 Ethics statement

Clinical data were collected retrospectively from electronic health records. This study, which was conducted in accordance with the Declaration of Helsinki, was approved by the institutional review board (IRB) at Shigei Medical Research Hospital (#20161219-1) and Kobayashi Medical Clinic (#20190925), as a retrospective observational study. The endpoint of this study approved at IRBs was to construct a decision-making support system that can provide dosage directions that are equal to or better than those of physicians who control dosages to maintain hemoglobin (Hb) values within 10–12 g/dl: the criterion stated in the Japanese hemodialysis guideline.

2.1.2 Clinical data collection

Clinical data were collected at two hospitals where Japanese adult hemodialysis patients were receiving anemia control treatment by board-certified senior members of the Japanese Society for Dialysis Therapy. Data were collected at Shigei Medical Research Hospital (Hospital S) from January 2015 through May 2019 and at Kobayashi Medical Clinic (Hospital K) from November 2018 through September 2019. All clinical data were anonymized. At Hospital S, the $S_1$ and $S_2$ datasets were prepared. Dataset $S_1$ was used for training the neural network; $S_2$ was used for raw data validation. At Hospital K, dataset $K_1$ was prepared and used for raw data validation. At both hospitals S and K, directions by physicians at every hemodialysis occasion, which are every one or two weeks depending on the hospitals, were recorded in the form of UP, DOWN, or STAY because dosages for administration were directed in units of one ampoule under hospital regulations. The hemodialysis patients were 350 per year at Hospital S and 90 per year at Hospital K. The cases of mortality were 35 per year at Hospital S and 10 per year at Hospital K. Hospital K was selected to examine the applicability of AISACS at smaller hospitals.

The patient selection criteria were the following: maintenance hemodialysis, no concomitant inflammation (CRP<0.3 mg/dL), no infectious disease, and no present cancer. Moreover, the data collection period for each patient case was chosen to include as many UP and DOWN directions as possible in both training and validation groups. This period-selection criterion was used because data for maintenance hemodialysis patients in stable condition include larger numbers of STAY directions than either UP or DOWN directions, indicating that appropriate timings of UP and DOWN decisions are significant for patient care.

As a result obtained from data selection criteria described above, dataset $S_1$ with $N=130$, $W=6080$, and dataset $S_2$ with $N=81$, $W=1857$ were prepared from Hospital S, where $N$ and $W$ respectively represent the number of patients and hemodialysis occasions. Dataset $S_1$ was used for training the neural network, whereas $S_2$ was used for raw data validation. Dataset $K_1$ was prepared and used for raw data validation with $N=16$ and $W=298$.

Darbepoetin alfa and epoetin beta pegol were used as ESAs. The ISs were provided in the form of sodium ferrous citrate, ferrous fumarate, and saccharated ferric oxide (Supplemental Table A1). The target range was set as 10.0–12.0 g/dl at Hospital S according to the Japanese hemodialysis guideline. The Hb values were controlled by physicians within target ranges of 74% in $S_1$ and 73% in $S_2$ (Supplemental Table A2). Also, ESA-resistant patients were excluded. Therefore, the mean administered dosages of darbepoetin alfa were 20.2±10.1 µg/week in $S_1$, 18.8±14.1 µg/week in $S_2$ and 20.4±13.5 µg/week in $K_1$. The mean administered dosages of epoetin beta pegol were 26.1±8.9 µg/week in $S_1$, 36.0±15.7 µg/week in $S_2$, with no use in $K_1$ (Supplemental Table A3).

2.1.3 Inputs and outputs for machine learning

Four items of blood examination were regarded as neural network inputs: Hb; mean corpuscular volume (MCV); ferritin; and transferrin saturation (TSAT). These items, their trends, and histories of dosages for ESAs and ISs up until the previous administration occasion were used as input parameters. Finally, AISACS outputs probabilities for ternary directions in the form of UP, STAY, and DOWN in ESAs, and UP and STAY for binary directions in ISs, as shown in Figure 1. Ternary directions were not needed for ISs because the ISs were set to stop after 6 weeks, in accordance with hospital regulations.

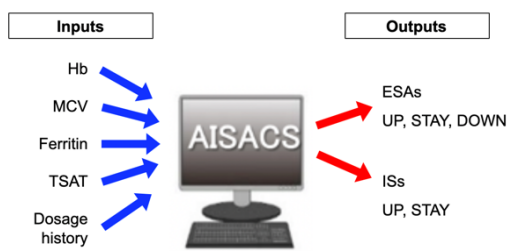

**Figure 1.** Inputs and outputs for machine learning.

2.1.4 Data rectification

One important difficulty in collecting administered dosage data is posed by "delayed decisions." For each hemodialysis occasion, patients underwent blood examinations. Usually the physicians then examined the results and gave administration directions. However, not all the decisions were made on the same day of the examination because of the delays in delivering the examination results to physicians caused by mechanical troubles, working time restrictions, and other factors. In such cases, the decision events were actually recorded with a week delay after the blood examination results on which the decision was actually based. Such a non-essential difference between blood examination and decision dates confused the neural network training process considerably. Therefore, we performed data rectification by moving the UP and DOWN decision dates to the exact dates on which the blood examinations were actually performed. This rectification procedure was done automatically and was confirmed by three physicians. The procedure was applied only for $S_1$ to be used for neural network training.

*2.2 Machine learning and validations*

2.2.1 Preliminary analyses

Before starting a deep learning approach, we applied simpler approaches to examine the complexity of our classification problem. Figure 2 portrays a principal component analysis (PCA) based on input data. From Fig. 2 using three principal components (PCs), it is apparent that almost all UP and DOWN decisions were readily classifiable using linear approaches, but UP and STAY, or STAY and DOWN are difficult to classify clearly using PCs. Moreover, several outliers exist, such as UP decisions located in the upper-right corner of Fig. 2(b). Based on these preliminary attempts, we decided to apply a deep learning approach, which is expected to work for such nonlinear, high-complexity classification problems.

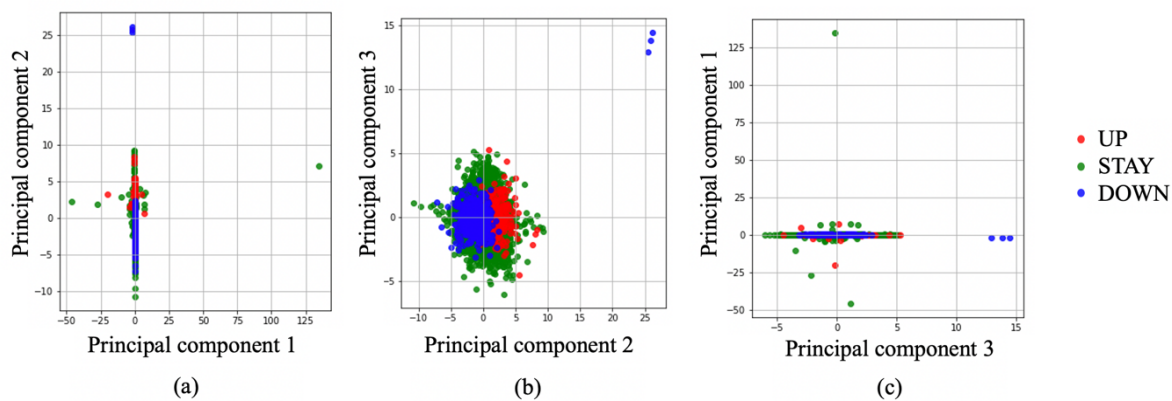

**Figure 2.** Classification by principal component analysis.

2.2.2 Machine learning setup

Machine learning codes were written using Keras with a TensorFlow backend [17,18]. The blood examination intervals for Ferritin/TSAT are usually longer than that of Hb/MCV. Therefore, we used independent neural networks of two kinds for the two forms of medication. Indeed, Hb and MCV are examined every week, whereas Ferritin and TSAT are examined every month, which means that only a quarter of the dataset has actual measured values of Ferritin and TSAT to predict ISs. For this reason, whereas a dense neural network was used for ESAs, a recurrent neural network (RNN) [19] was used for ISs as a more effective method when fewer data are available. Considering the tradeoff between training data size and representation ability, a recursive layer with sequence size two was added to the dense neural network, so two successive timings are passed as inputs. Both networks used 10 hidden layers with $L^1$ regularization and drop-out techniques [22] to prevent overfitting phenomena. Other training parameters and hardware used for machine learning are presented in Table 4A.

2.2.3 Validations

We defined correct classification rates $R_{TOTAL}$ as

$$R_{TOTAL} = \frac{\text{number of correct decisions}}{\text{number of input decision data}},$$

which were the ratios by which AISACS gave the same directions on the same dates as those given by physicians. We also defined $R_{UP}$, $R_{STAY}$, and $R_{DOWN}$ by confining the decision to each class. Using these values, we performed the following validations of two types.

- "Leave one patient out" cross-validation (LOPO)
  LOPO was performed by removing data of one patient from the dataset. The neural network was trained using the remaining $N$-1 patient data. Then the removed patient data were used to evaluate the performance of the trained neural network. After repeating this procedure $N$ times, correct classification rates were calculated using $N$ patients results. The $S_1$ dataset was used for LOPO.

- Raw data validation (RDV)
  RDV was performed using $S_2$ and $K_1$. First, we trained the neural network using $S_1$. Then the correct classification rates were calculated using $S_2$ (RDV_S) and $K_1$ (RDV_K). Training and validation processes are completely independent in RDV_S and RDV_K.

Validations performed in this study are presented in Table 1 and are shown schematically in Figure 3.

Table 1. Validations and datasets using $S_1$ and $S_2$ from Hospital S and $K_1$ from Hospital K.

| Name | Validation procedure | Dataset for training | Dataset for validation |
|---|---|---|---|
| LOPO | Leave one patient out cross-validation | $S_1$ | |
| RDV_S | Raw data validation | $S_1$ | $S_2$ |
| RDV_K | Raw data validation | $S_1$ | $K_1$ |

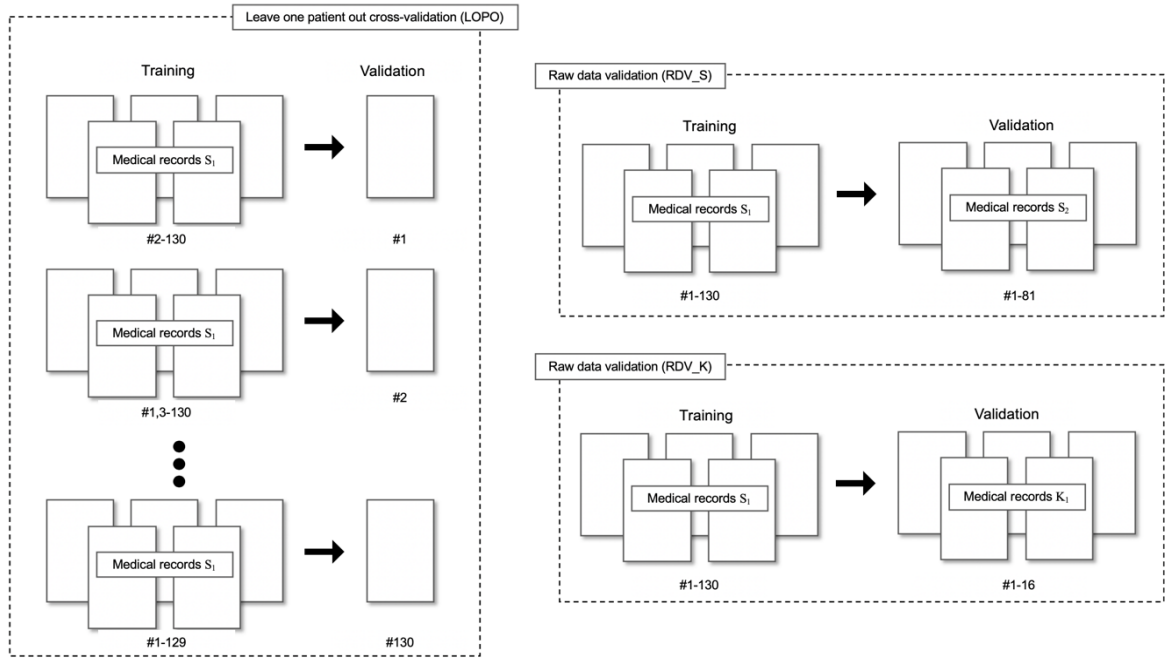

**Figure 3.** Leave one patient out (LOPO) cross-validation and raw data validations (RDV) procedures.

2.2.4 Class-imbalanced training data

Although we selected the clinical data period that includes plentiful UPs and DOWNs, the numbers of different directions included in the dataset are still markedly imbalanced. For example, in dataset $S_1$, ESA directions by physicians comprised 344 UPs, 585 DOWNs, and 5151 STAYs. Simple machine learning using such an imbalanced dataset led to AI always outputting the STAY direction to achieve the highest $R_{TOTAL}$. However, the timings of UP and DOWN are much more important for the present problem. Such a discrepancy can usually be controlled by class weights, respectively strengthening and weakening the effects of minority and majority classes on the target functions. Although values of class weights are usually defined using the inverse ratios of quantities of data, class-imbalance was not improved sufficiently for AISACS. Therefore, they were further adjusted to strengthen minority classes by trial and error so that $R_{UP}$, $R_{STAY}$, and $R_{DOWN}$ are approximately equal in $S_1$.

2.2.5 Two-step classification for the ternary classification for ESAs

Because the ESA administration belongs to ternary classification problems, three probability values of $P_{UP}$, $P_{STAY}$, and $P_{DOWN}$, respectively corresponding to UP, STAY, and DOWN directions, were computed as outputs from the neural network. The simplest method for classification is to adopt a direction that gives the highest probability value. However, such a simple algorithm does not seem to work for the present situation in which the timings of UP and DOWN are crucially important to appropriate anemia control. Therefore, we propose the following procedure for the ternary classification problem: First, we set a threshold value $T$. The direction is assigned as STAY if the probability of STAY was larger than $T$. Otherwise, UP or DOWN, which has a larger probability, is assigned, as portrayed in Fig. 4. We designate the union of UP and DOWN classes as NON-STAY in the following sections.

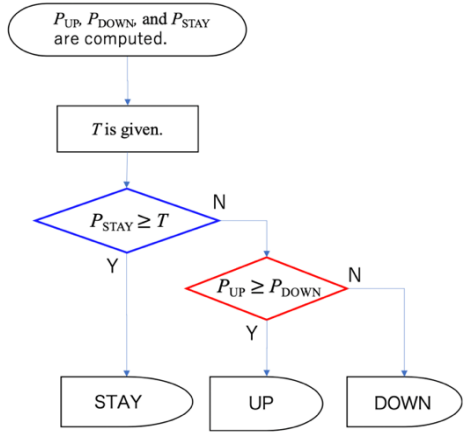

**Figure 4.** Flow chart of two-step classification for ESAs.

## 3. Results

### 3.1 Classification between STAY and NON-STAY directions

As described in 2.2.1, assigning classification for ESA administration between STAY and NON-STAY directions is much more difficult than assigning classification between UP and DOWN directions. Therefore, we examined the classification ability of AISCAS by drawing receiver operating characteristic (ROC) curves for STAY and NON-STAY directions by changing the threshold T. Figure 5 portrays ROC curves and area under curve (AUC) values for ESAs and ISs. Threshold T is varied from 0 to 1. For ESAs, RDV_K shows lower AUC than RDV_S, which can be a consequence of the fact that AISCAS was trained using data from Hospital S. This point is discussed in Section 4.

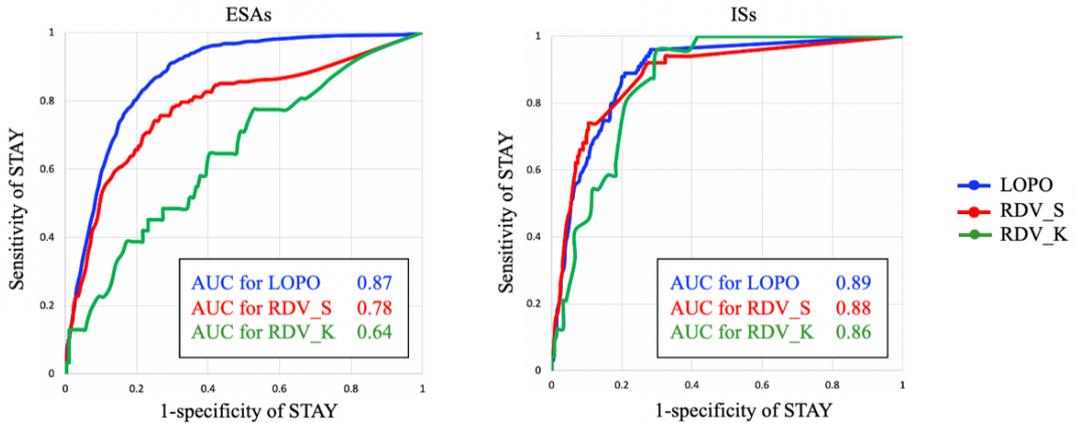

**Figure 5.** ROC curves and AUC values for ESAs and ISs.

### 3.2 Correct classification rates after fixing threshold T

On actual situations in hospitals, a threshold value T discussed in 2.2.4 should be decided. One possible strategy using the ROC curves is to choose T corresponding to the nearest point on the ROC curve from point (x, y) = (0, 1) to achieve similar abilities for both STAY and NON-STAY. For dataset $S_1$, this value appeared to be 0.475 for ESAs and 0.470 for ISs, which we adopted also for validations and which gives the correct classification rates $R_{TOTAL}$ for LOPO, RDV_S, and RDV_K as 80%, 77%, and 72% for ESAs and 81%, 87%, and 80% for ISs.

*3.3 Examining incorrect classification cases*

To analyze reasons for incorrect classification cases, we reviewed them carefully one-by-one, which revealed some directions by AISACS that appeared to be appropriate from a medical perspective, even though they differed from the physician's recorded directions. We defined these as "clinically appropriate" directions. Moreover, we found that a characteristic type exists in "clinically appropriate" directions, which we defined as a "before physician" direction. In "before physician" directions, AISACS gave the same UP or DOWN directions with physicians, but gave it a week or so earlier than the physician did. "Before physician" directions are calculable automatically by counting up to three earlier administration occasions than the physician. Although such "before physician" directions are counted as incorrect classifications in 3.2, they portray an interesting feature of AISACS. Other "clinically appropriate" directions are the other portion in clinically appropriate directions judged by board-certified doctors. The rate of "before physician" in validations LOPO, RDV_S, and RDV_K were, respectively, 9%, 7%, and 8% for ESAs and 5%, 5%, and 5% for ISs. The rate of "clinically appropriate: other" directions were, respectively, 8%, 8%, and 15% for ESAs and 9%, 6%, and 10% for ISs. Ratios for "correct classification," "clinically appropriate: before physician," and "clinically appropriate: other" are shown respectively in Figs. 6 and 7.

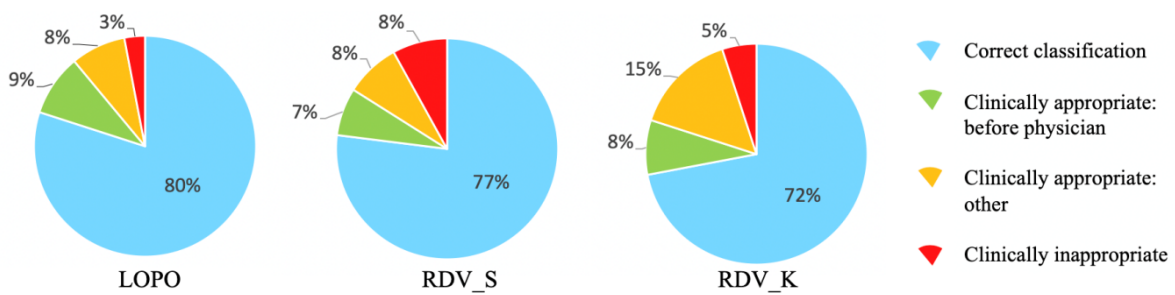

**Figure 6.** Categorization of classification results by AISACS for ESAs with $T$=0.475.

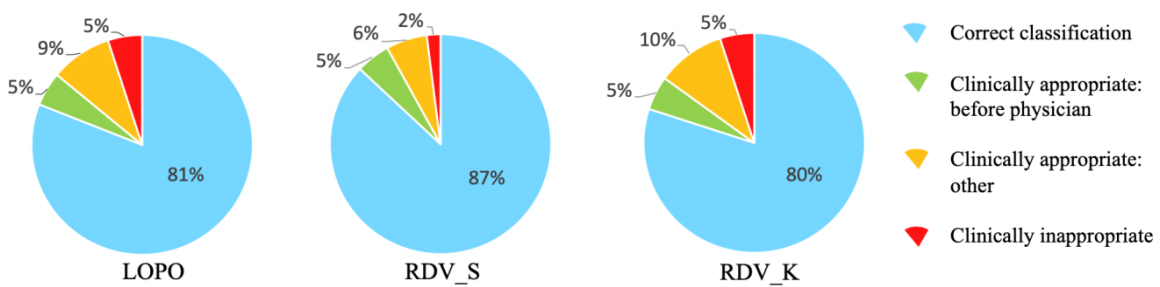

**Figure 7.** Categorization of classification results by AISACS for ISs with $T$=0.470.

Finally, gross rates of appropriate directions, which were the sum of "correct classification," "clinically appropriate: before physician," and "clinically appropriate: other," in validations LOPO, RDV_S, and RDV_K were 97%, 92%, and 95% for ESAs and 95%, 98%, and 95% for ISs.

## 4. Discussion

Four features of AISACS are particularly important. The first feature is what AI learns: reactions of living bodies or decisions of experienced physicians. Systems for predicting future Hb values of maintenance hemodialysis patients using AI technology have been reported as described in Section 1. We adopted a different approach by which AI learns from experienced physicians' dosage directions. Actually, experienced physicians do not calculate detailed values of vital reactions when deciding dosages. We selected five items of blood examination, their trends, and dosage histories as inputs by looking at the judgments reported by physicians.

A second feature is proper data selection and rectification. For example, "delayed decisions" appear frequently in real datasets because of mechanical difficulties and working time restrictions. In such cases, the decision dates were recorded with a one or two week lag after the blood examination actually occurred. Such a nonessential difference between blood examination and actual decision dates confuse the training process of our neural network considerably. Therefore, we moved the dates of UP and DOWN directions to dates on which the decisions were actually based. Such a data rectification procedure functioned well to make the training process efficient, even though the training in this study was based on a small sample of data. Figure 8 presents correct classification rates for ESAs in $S_1$ improved during AISACS development: in (a) with a few layers in a neural network with no weighting techniques, it almost always yields the STAY direction. Then, by a tuning of class weights, the correct classification rates $R_{UP}$, $R_{STAY}$, and $R_{DOWN}$ became approximately equal to each other as portrayed in Fig. 8(b). By increasing the number of layers and by adding several means from (c)–(e) such as class weights, dosage histories reference and two-step classification, the correct classification rates, especially for UP and DOWN, were improved considerably.

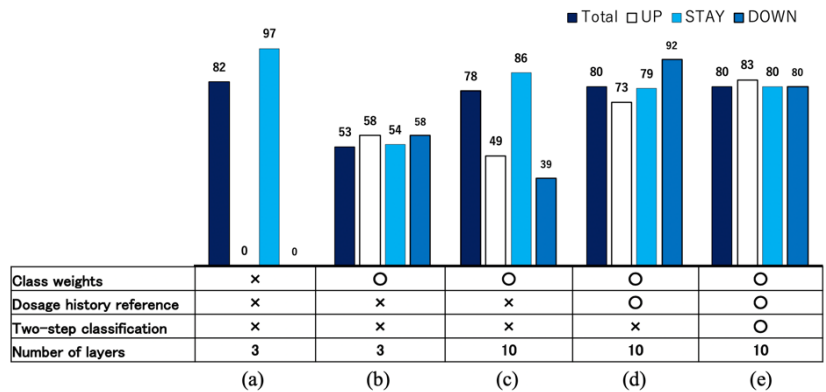

**Figure 8.** Correct classification rates for ESAs in $S_1$ during developing AISACS.

When comparing the AUCs in raw data validation using data from hospitals S and K (RDV_S vs. RDV_K), the AUC from RDV_S was found to be higher than that from RDV_K because AISACS was trained using the dataset from Hospital S. Apparently, AISACS has some affinity to physicians at Hospital S. However, the "clinically appropriate" rates for Hospital K were sufficient, which suggests that AISACS has a certain degree of flexibility.

A third feature is the multi-class classification for ESAs. The direction timings of UP and DOWN are crucially important for appropriate anemia control. Therefore, we set a threshold between STAY and NON-STAY directions using the ROC curve based on probabilities calculated using the neural network. Then, NON-STAY is classified to UP or DOWN simply by comparison of their probabilities. It is possible to tune the frequency of decision changes by adjusting the threshold value. For example, if the threshold were set at a higher value, then AISACS would give more frequent UPs and DOWNs. This feature might be useful when AISACS is applied at different hospitals.

A fourth feature is that AISACS sometimes shows better timing than physicians for changing dosage directions as described in Section 3.3. The appearance of "before physician" directions portrays an interesting feature of AISACS, which can contribute to helping physicians to see right timings to increase or decrease dosages. There is an additional interesting point here. As presented in Section 3.1, the AUC value from RDV_K for ESAs was quite lower than that from RDV_S, which might be attributable to AISACS learned decisions of physicians at Hospital S. However, many of the incorrect classification cases were regarded as clinically correct through multiple doctors' reviews. Actually, on one hand, the AUC of RDV_K for ESAs is the lowest among four raw data validations. On the other hand, the "clinically appropriate decision" portion of it was the highest.

The present study has the following limitations. We conducted retrospective analysis for patients from only two hospitals, involving only Japanese patients with a small sample size. Moreover, we did not evaluate the cost of ESAs and irregular cases such as patients with conditions aggravated by other diseases. Considering the endpoint approved at IRBs for this study, it is difficult at the moment, to ascertain whether AISACS can give better directions than physicians, or not. A prospective, multi-center study is therefore needed, especially for confirmation of patient safety.

## 5. Conclusions

Preventing anemia is important to improve the prognosis and quality of life of hemodialysis patients. However, the pathophysiology associated with anemia is complicated. It requires a great deal of experience to control anemia cases adequately. The number of such physicians is insufficient. For this reason, we have constructed AISACS. The challenges and contributions to anemia control practices described in this paper are the following.

- Not-so-large training dataset: We have constructed proper data selection and rectification procedures that play important roles in enhancing machine learning efficiency with small datasets.
- Importance of appropriate timing of dosage changes: AISACS provides ternary directions for ESAs equipped with a threshold value to control NON-STAY and STAY decision tendencies.
- Widely diverse health conditions of dialysis patients: Patients have several legal and economic constraints. A feature that is unique to AISACS is that it learns dosage directions from physicians using no prediction model based on biochemistry or physiology.

In addition, an interesting feature of AISACS is that it sometimes produces "clinically appropriate" directions that are different from those of physicians, but which are nonetheless proper. Finally, AISACS has achieved a quite high gross rates of correct classification, which means giving the same direction with physicians on the same date, as 72%–87% and clinically appropriate classification, although it includes different decisions from those of physicians as 92%–98% through several validations. These results attest to AISACS' promising possibilities for clinical applications after wider validation through a prospective, multi-center study.

**Author Contributions:** T.O., H.I., and H.S developed the concept and designed the research; T.O., H.I., Y.S., H.S., V.H., and S.H. conducted the research and acquired data; T.O., H.I, Y.S., H.S., M.K., S.H., and K.S. analyzed and interpreted data; T.O. and H.S. wrote and reviewed the manuscript.

**Funding:** This work was partially supported by JST CREST Grant Number JPMJCR15D1, Japan. This work was also partially supported by a Grant for Promotion of Science and Technology in Okayama Prefecture by MEXT, Japan

**Conflicts of Interest:** The authors declare that none has any competing interest.

# Appendix A

**Table A1.** Background

|  | Hospital S | | Hospital K |
| --- | --- | --- | --- |
|  | $S_1$ | $S_2$ | $K_1$ |
|  | ($N$=130) | ($N$=81) | ($N$=16) |
| Sex | | | |
|   Male | 78 | 49 | 9 |
|   Female | 52 | 32 | 7 |
| Age (years) | | | |
|   Mean | 78.0 | 65.6 | 68.3 |
|   Range | 26–89 | 35–84 | 51-86 |
| Hemodialysis period | | | |
|   Mean | 9.7 | 11.9 | 3.5 |
|   Range | 1–42 | 1–35 | 1-11 |
| Primary disease | | | |
|   Chronic glomerulonephritis (CGN) | 62 | 49 | 2 |
|   Diabetes | 32 | 19 | 8 |
|   Renal sclerosis | 15 | 10 | 3 |
|   Cystic kidney | 8 | 0 | 0 |
|   Ureteral stone | 1 | 0 | 0 |
|   Other | 12 | 3 | 3 |
| Erythropoiesis-stimulating agent | | | |
|   Darbepoetin alfa | 121 | 68 | 16 |
|   Epoetin beta pegol | 9 | 11 | 0 |
|   No administration | 0 | 2 | 0 |
| Iron supplement | | | |
|   Sodium ferrous citrate | 32 | 22 | 1 |
|   Ferrous fumarate | 5 | 4 | 0 |
|   Saccharated ferric oxide | 50 | 26 | 11 |
|   Sodium ferrous citrate + Saccharated ferric oxide | 8 | 7 | 0 |
|   No administration | 35 | 22 | 4 |

**Table A2.** Ranges of blood examination dataset

| Dataset | Examination item | Start | End |
| --- | --- | --- | --- |

| | | | | |
|---|---|---|---|---|
| Hospital S | S₁ (N=130) | Hb (g/dl) | 10.3±0.8 | 10.8±0.6 |
| | | MCV (fl) | 89.6±7.3 | 92.5±5.3 |
| | | TSAT (%) | 21.6±11.4 | 31.7±29.6 |
| | | Ferritin (ng/ml) | 53.7±50.3 | 72.7±46.7 |
| | S₂ (N=81) | Hb (g/dl) | 10.5±0.8 | 10.9±0.9 |
| | | MCV (fl) | 92.0±7.0 | 92.6±6.1 |
| | | TSAT (%) | 21.7±10.0 | 26.8±14.1 |
| | | Ferritin (ng/ml) | 52.6±37.8 | 61.8±41.3 |
| Hospital K | K₁ (N=16) | Hb (g/dl) | 11.1±0.7 | 10.9±0.7 |
| | | MCV (fl) | 93.8±5.7 | 95.3±4.5 |
| | | TSAT (%) | 23.2±10.1 | 23.4±6.4 |
| | | Ferritin (ng/ml) | 50.1±31.9 | 34.8±14.0 |

**Table A3.** Mean ESAs administered dosage

| | | Darbepoetin alfa | Epoetin beta pegol |
|---|---|---|---|
| Hospital S | S₁ (N=130) | 20.2±10.1 μg/week (N=120) | 26.1±8.9 μg/2 weeks (N=10) |
| | S₂ (N=81) | 18.8±14.1 μg/week (N=68) | 36.0±15.7 μg/2 weeks (N=11) |

**Table A4.** Training parameters and a hardware for machine learning

| Parameter | Setting |
|---|---|
| Number of units in each layer | 512 |
| Drop-out | 20% |
| Regularization coefficient | 0.3 |
| Optimization method | Adam method |
| Number of inputs | 16 |
| Number of outputs | ESAs, 3; ISs, 2 |
| Number of epochs | 1000 |
| Hardware | iMac with CPU: Intel4I Core I7, RAM: 32GB, GPU: AMD Radeon R9 M395X |
| CPU time for one training | Approximately 600 s |

**References**

1. Sato, Y.; Fujimoto, S.; Konta, T.; Iseki, K.; Moriyama, T.; Yamagata, K.; Tsuruya, K.; Narita, I.; Kondo, M.; Kasahara, M. et al. Anemia as a risk factor for all-cause mortality: obscure synergic effect of chronic kidney disease. *Clin. Exp. Nephrol.* **2018**, *22*, 388-394, doi:10.1007/s10157-017-1468-8.
2. Zhao, L.; Hu, C.; Cheng, J.; Zhang, P.; Jiang, H.; Chen, J. Haemoglobin variability and all-cause mortality in haemodialysis patients: A systematic review and meta-analysis. *Nephrology (Carlton)* **2019**, *24*, 1265-1272, doi:10.1111/nep.13560.
3. Hanafusa, N.; Nakai, S.; Iseki, K.; Tsubakihara, Y. Japanese society for dialysis therapy renal data registry – a window through which we can view the details of Japanese dialysis population. *Kidney Int. Suppl (2011)* **2015**, *5*, 15-22, doi:10.1038/kisup.2015.5.
4. Ortiz, A.; Sanchez-Nino, M.D.; Crespo-Barrio, M.; De-Sequera-Ortiz, P.; Fernandez-Giraldez, E.; Garcia-Maset, R.; Macia-Heras, M.; Perez-Fontan, M.; Rodriguez-Portillo, M.; Salgueira-Lazo, M. et al. The Spanish Society of Nephrology (SENEFRO) commentary to the Spain GBD 2016 report: Keeping chronic kidney disease out of sight of health authorities will only magnify the problem. *Nefrologia* **2019**, *39*, 29-34, doi:10.1016/j.nefro.2018.09.002.
5. Swaminathan, S.; Mor, V.; Mehrotra, R.; Trivedi, A.N. Effect of Medicare dialysis payment reform on use of erythropoiesis stimulating agents. *Health Serv Res* **2015**, *50*, 790-808, doi:10.1111/1475-6773.12252.
6. Bello, A.K.; Levin, A.; Tonelli, M.; Okpechi, I.G.; Feehally, J.; Harris, D.; Jindal, K.; Salako, B.L.; Rateb, A.; Osman, M.A. et al. Assessment of Global Kidney Health Care Status. *JAMA* **2017**, *317*, 1864-1881, doi:10.1001/jama.2017.4046.
7. Thongprayoon, C.; Kaewput, W.; Kovvuru, K.; Hansrivijit, P.; Kanduri, S.R.; Bathini, T.; Chewcharat, A.; Leeaphorn, N.; Gonzalez-Suarez, M.L.; Cheungpasitporn, W. Promises of Big Data and Artificial Intelligence in Nephrology and Transplantation. *J Clin Med* **2020**, *9*, doi:10.3390/jcm9041107.
8. Xie, G.; Chen, T.; Li, Y.; Chen, T.; Li, X.; Liu, Z. Artificial Intelligence in Nephrology: How Can Artificial Intelligence Augment Nephrologists' Intelligence? *Kidney Dis (Basel)* **2020**, *6*, 1-6, doi:10.1159/000504600.
9. Goldstein, B.A.; Chang, T.I.; Mitani, A.A.; Assimes, T.L.; Winkelmayer, W.C. Near-term prediction of sudden cardiac death in older hemodialysis patients using electronic health records. *Clin. J. Am. Soc. Nephrol.* **2014**, *9*, 82-91, doi:10.2215/CJN.03050313.
10. Gaweda, A.E.; Aronoff, G.R.; Jacobs, A.A.; Rai, S.N.; Brier, M.E. Individualized anemia management reduces hemoglobin variability in hemodialysis patients. *J. Am. Soc. Nephrol.* **2014**, *25*, 159-166, doi:10.1681/ASN.2013010089.
11. Gaweda, A.E.; Jacobs, A.A.; Aronoff, G.R.; Brier, M.E. Individualized anemia management in a dialysis facility – long-term utility as a single-center quality improvement experience. *Clin Nephrol* **2018**, *90*, 276-285, doi:10.5414/CN109499.
12. Rogg, S.; Fuertinger, D.H.; Volkwein, S.; Kappel, F.; Kotanko, P. Optimal EPO dosing in hemodialysis patients using a non-linear model predictive control approach. *J Math Biol* **2019**, *79*, 2281-2313, doi:10.1007/s00285-019-01429-1.

13. Escandell-Montero, P.; Chermisi, M.; Martinez-Martinez, J.M.; Gomez-Sanchis, J.; Barbieri, C.; Soria-Olivas, E.; Mari, F.; Vila-Frances, J.; Stopper, A.; Gatti, E. et al. Optimization of anemia treatment in hemodialysis patients via reinforcement learning. *Artif Intell Med* **2014**, *62*, 47-60, doi:10.1016/j.artmed.2014.07.004.
14. Barbieri, C.; Bolzoni, E.; Mari, F.; Cattinelli, I.; Bellocchio, F.; Martin, J.D.; Amato, C.; Stopper, A.; Gatti, E.; Macdougall, I.C. et al. Performance of a Predictive Model for Long-Term Hemoglobin Response to Darbepoetin and Iron Administration in a Large Cohort of Hemodialysis Patients. *PLoS One* **2016**, *11*, e0148938, doi:10.1371/journal.pone.0148938.
15. Barbieri, C.; Molina, M.; Ponce, P.; Tothova, M.; Cattinelli, I.; Ion Titapiccolo, J.; Mari, F.; Amato, C.; Leipold, F.; Wehmeyer, W. et al. An international observational study suggests that artificial intelligence for clinical decision support optimizes anemia management in hemodialysis patients. *Kidney Int.* **2016**, *90*, 422-429, doi:10.1016/j.kint.2016.03.036.
16. Bucalo, M.L.; Barbieri, C.; Roca, S.; Ion Titapiccolo, J.; Ros Romero, M.S.; Ramos, R.; Albaladejo, M.; Manzano, D.; Mari, F.; Molina, M. The anaemia control model: Does it help nephrologists in therapeutic decision-making in the management of anaemia? *Nefrologia* **2018**, *38*, 491-502, doi:10.1016/j.nefro.2018.03.004.
17. Chollet François, A.O. Keras. Available online: https://keras.io (accessed on 6 August, 2020).
18. Abadi, M.; Agarwal, A.; Barham, P.; Brevdo, E.; Chen, Z.; Citro, C.; Corrado, G.S.; Davis, A.; Dean, J.; Devin, M. et al. TensorFlow: Large-scale machine learning on heterogeneous systems. Available online: https://www.tensorflow.org/ (accessed on 6 August, 2020).
19. Hochreiter, S.; Schmidhuber, J. Long short-term memory. *Neural Comput.* **1997**, *9*, 1735-1780, doi:10.1162/neco.1997.9.8.1735.
20. Bishop, C.M. *Pattern recognition and machine learning*, Springer: 2006.